\documentclass[letterpaper, 10 pt, conference]{ieeeconf}  
\usepackage[letterpaper, left=0.667in, right=0.667in, bottom=0.75in, top=0.792in]{geometry}

\IEEEoverridecommandlockouts                              

\usepackage{graphicx} 
\usepackage{amsmath} 
\usepackage{amssymb}  
\usepackage{cite}
\usepackage{caption}
\usepackage{enumerate}
\usepackage{subcaption}
\usepackage{hyperref}
\usepackage[linesnumbered,ruled]{algorithm2e}
\usepackage{multirow}
\usepackage{makecell}
\usepackage{hhline}

\usepackage{color}
\newcommand{\norm}[1]{\left\lVert#1\right\rVert}

\title{\LARGE \bf
Reconstruction of Backbone Curves for Snake Robots
}

\author{Tianyu Wang$^{1,*}$, Bo Lin$^{2,*}$, Baxi Chong$^{2}$, Julian Whitman$^{1}$, Matthew Travers$^{1}$, \\Daniel I. Goldman$^{2}$, Greg Blekherman$^{2}$, Howie Choset$^{1}$
\thanks{*These authors contributed equally}
\thanks{$^{1}$Tianyu Wang, Julian Whitman, Matthew Travers and Howie Choset are with Carnegie Mellon University, Pittsburgh, PA 15213, USA.  {\{\tt\small tianyuw2, jwhitman, mtravers, choset\}@andrew.cmu.edu}}
\thanks{$^{2}$Bo Lin, Baxi Chong, Daniel I. Goldman and Greg Blekherman are with Georgia Institute of Technology, Atlanta, GA 30332, USA. {\{\tt\small bo.lin@math., bchong9@, daniel.goldman@physics., greg@math.\}gatech.edu}}
}

\begin{document}
\maketitle
\thispagestyle{empty}
\pagestyle{empty}

\begin{abstract}
Snake robots composed of alternating single-axis pitch and yaw joints have many internal degrees of freedom, which make them capable of versatile three-dimensional locomotion.
In motion planning process, snake robot motions are often designed kinematically by a chronological sequence of continuous backbone curves that capture desired macroscopic shapes of the robot.
However, as the geometric arrangement of single-axis rotary joints creates constraints on the rotations in the robot, it is challenging for the robot to reconstruct an arbitrary 3D curve.
When the robot configuration does not accurately achieve the desired shapes defined by these backbone curves, the robot can have unexpected contacts with the environment, such that the robot does not achieve the desired motion.
In this work, we propose a method for snake robots to reconstruct desired backbone curves by posing an optimization problem that exploits the robot's geometric structure.
We verified that our method enables fast and accurate curve-configuration conversions through its applications to commonly used 3D gaits.
We also demonstrated via robot experiments that 1) our method results in smooth locomotion on the robot; 2) our method allows the robot to approach the numerically predicted locomotive performance of a sequence of continuous backbone curve.
\end{abstract}

\section{Introduction}

\begin{figure}[t]
\centering
\includegraphics[width=0.9\columnwidth]{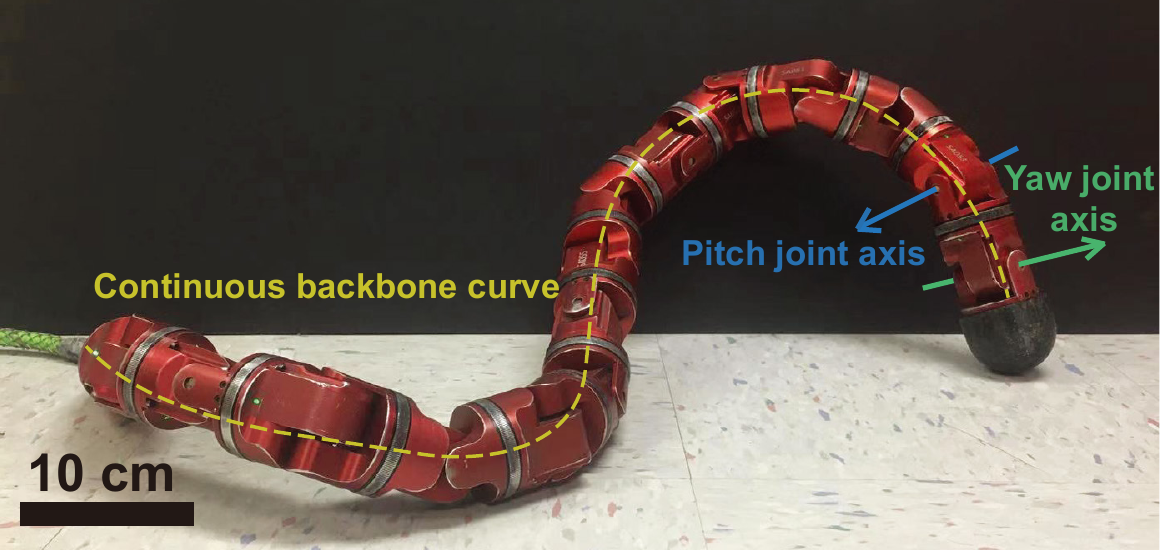}
\caption{An example of a ``twist-free" snake robot design with alternating single-axis pitch and yaw joints.
}
\vspace{-1.5em}
\label{fig:robot}
\end{figure}

Snake robots are a class of hyper-redundant mechanisms capable of achieving different types of locomotion by coordinated flexing of their bodies. 
One of the well-established snake robot designs is composed of alternating one degree of freedom (DOF) pitch and yaw bending joints (as shown in Fig. \ref{fig:robot}), which allows 3D versatile motion \cite{wright2007design,takaoka2011snake,transeth2008snake,fu2020lateral}. 
Such a robot design is also called ``twist-free'' since it does not have direct actuation of the twist (rotation about the longitudinal axis of the body) DOF \cite{nilsson1998snake}.
Inspired by the shapes of biological snakes with many vertebrae, finite-length continuous \textit{backbone curves} are designed to capture desired macroscopic shapes of robots \cite{hirose1993biologically,chirikjian1994modal}.
Often, the motion of a snake robot is planned kinematically by a chronological sequence of backbone curves \cite{hatton2010generating,takemori2018gait}.
Once properly designed, these sequences of backbone curves have been shown by prior work to generate effective, biologically-inspired, locomotion such as lateral undulation, sidewinding, and sinus lifting \cite{hirose1993biologically,burdick1993sidewinding,ma2004dynamic}.
In order to replicate those motions on the physical robot, we have to match the shape of a robot made up of discrete segments to the continuous curves.
These backbone curves lie in 3D space, but the geometric arrangement of single-axis rotary joints creates constraints on the rotations in the robot, making this shape-matching problem challenging for twist-free snake robots. 
When the body shape does not match the desired backbone curve, the desired robot-environment contacts are not achieved. The robot may then have undesired contacts with the environment, impeding locomotion.
This paper presents a method for twist-free snake robots to accurately reconstruct desired 3D backbone curves via a constrained optimization problem. This enables the robot to locomote effectively by following a sequence of backbone curves.

In prior work, one widely employed method to reconstruct 3D continuous backbone curves with snake robot configurations is to decompose the 3D curve into 2D sub-curves and separately prescribe the pitch and yaw joint angles by parameterized sinusoidal functions according to these sub-curves \cite{tesch2009parameterized, astley2015modulation, zhong2020frequency}. 
This method simplifies the conversion process and is easily implemented. 
However it neglects the intrinsic twisting properties in the given 3D curve \cite{o2014elementary}, and predefined sinusoidal functions often cannot provide good approximations to sub-curves, which can lead to discrepancies between the resultant robot configuration and the desired backbone curve.
Another branch of approaches to the reconstruction problem is to discretize the curve by fitting piece-wise linear segments to the continuous finite-length curve with optimization tools \cite{chirikjian1995kinematics, mochiyama1999shape, andersson2008discretization}.
However, such approaches often assume two or more DOFs of rotation capability at each joint, preventing their direct application to the alternating-rotation robot geometry.
A method introduced in \cite{hatton2010generating} unified these two types of approaches by first fitting the robot configurations to the curves with the optimization algorithm, then smoothing out the fitted joint angles over time with parameterized sinusoidal functions.  
This optimization-based algorithm allows conversions from individual 3D backbone curves to individual robot configurations, but the output joint angles are not continuous over time so the smoothing step is necessary.
However, the smoothing process does not directly consider the desired backbone curve, such that the joint angles after smoothing may no longer fit the given curves.

In this work, we propose a method for snake robots with an alternating single-axis pitch and yaw joints design to reconstruct desired 3D backbone curves.
We develop the constrained optimization algorithm in such a way that it exploits the geometric structure of the robot.
We test our reconstruction approach on 3D backbone curves common in the literature \cite{hirose1993biologically, astley2015modulation, tanaka2009study, zhen2015modeling}.
By comparing individual robot configurations with the given backbone curves, we show that our reconstruction method allows the robot configurations to match the given curves more precisely than the state-of-the-art method presented in previous literature \cite{tesch2009parameterized,astley2015modulation, zhong2020frequency,hatton2010generating}. 
We also conduct physical robot experiments where we apply our method in real-time while executing 3D gaits. 
We experimentally validate that our method ensures that small changes in the desired backbone curves result in small changes in the robot configuration, thus the commanded joint angles vary smoothly.
We then demonstrate that our method allows the robot to approach the numerically predicted motion of a sequence of desired continuous backbone curves.

\vspace{-0.5em}
\section{Related work} \label{sec:related_work}
There are two primary approaches in prior work for reconstruction of the 3D continuous backbone curves with discrete robot configurations in snake robot locomotion.

The first dominant approach to the configuration reconstruction problem is to decompose the 3D curve into 2D sub-curves and derive pitch and yaw joint angles separately with simple heuristics.
Widely employed in gait design and control for snake robot with alternating single-axis pitch and yaw joints \cite{tesch2009parameterized, astley2015modulation, zhong2020frequency}, this approach first decomposes the desired 3D backbone curves to two 2D sub-curves in the vertical plane and the horizontal plane, and fits the vertical and horizontal body shape of the robot to the sub-curves by prescribing the pitch and yaw joint angles with two parameterized sinusoidal equations.
This approach is simple to implement, but, during the decomposition process the twisting properties of the desired curves are often neglected, and predefined parameterized sinusoidal equations often cannot provide good approximations to sub-curves.
Thus, the given curves cannot be fully replicated, leading to errors between the robot configurations and the desired curves.

The other dominant approach to the reconstruction problem is discretization of the 3D curve by fitting piece-wise linear segments to the continuous finite-length curve with optimization tools.
Chirikjian and Burdick \cite{chirikjian1995kinematics} proposed modal functions for describing arbitrary 3D backbone curves and developed algorithms for fitting discrete serial mechanisms to desired curves.
Mochiyama et al. \cite{mochiyama1999shape} considered the discretization problem as a shape-tracking control problem by deriving a control law for the joints such that the robot converges to the desired shape.
Anderson \cite{andersson2008discretization} derived analytical expressions for the relative orientations of the links in the configuration with a bisection search method.
However, these approaches assume a mechanism with universal joints, i.e., at least two intersecting rotation axes at each joint, making it challenging to directly implement these approaches on the alternating-rotation robot geometry.

Hatton and Choset \cite{hatton2010generating} proposed the state-of-the-art method to solve the reconstruction problem for the snake robots with alternating-rotation geometry by combining the two major types of approaches, outperforming previous approaches in reconstruction accuracy.
First, an optimization problem was formulated that fixed one endpoint of the robot and consecutively optimized joint angles to ``pull" the rest of links towards the desired backbone curve.
Solving this optimization problem produces conversions from individual 3D backbone curves to individual robot configurations.
However, when applied to a chronological sequence of the desired backbone curve, the joint angles outputs were not smooth over time, which cannot produce a continuous motion on the robot.
To address the discontinuity in the optimizer outputs, a post-processing smoothing step was used in which, parameterized sinusoidal equations were fit to the trajectories that the optimization algorithm produced.
However, since the smoothing process does not directly consider the desired backbone curve, errors often emerge in the process of fitting the joint angle trajectories to sinusoidal functions, which can result in shapes that no longer fit the desired backbone curves well.

\section{Methods} \label{sec:method}
We address the problem of converting from a 3D continuous backbone curve to an alternating-pitch-and-yaw robot configuration by posing an optimization problem that exploits the geometric structure of the robot.
We divide our reconstruction strategy into three steps: 1) sampling the desired joint positions in the robot's work-space based on the given backbone curve, 2) deriving the constraints based on the robot's geometric structure, designing the objective function, and running an optimization algorithm to obtain optimal joint positions in work-space, and 3) translating the optimal joint positions in work-space to joint angles in joint space that can be executed by the robot.

\vspace{-0.5em}
\subsection{Sampling the backbone curve}
In gait design or motion planning for a snake robot, a finite-length continuous backbone curve defines the robot body's desired shape. 
To simplify the optimization process without losing the shape properties of the given 3D backbone curve, we sample a set of points from the curve.
The 3D Cartesian coordinates of the samples are denoted by $C_0, C_1, ..., C_{n+1}$, where $n$ is the number of joints on the robot and $C_i = [C_{i,x}, C_{i,y}, C_{i,z}]^T \in \mathbb{R}^3$.
$C_0$ and $C_{n+1}$ are the anterior and posterior endpoints of the curve, and $C_1, ..., C_n$ evenly divide the curve into $n+1$ segments with the same arc length.
Thus, samples $C_i$ serve as the desired positions for the robot joints ($C_1$ to $C_{n}$) and anterior and posterior endpoints of the robot ($C_0$ and $C_{n+1}$).

\vspace{-0.5em}
\subsection{Robot joint position optimization}
We next construct the objective function for the optimization problem.
We use $n+2$ coordinates $C'_0, C'_1, ..., C'_{n+1}$ to denote the positions for the anterior endpoint of the robot, $n$ joints of the robot, and the posterior endpoint of the robot.
We can thus find the optimal robot configuration by minimizing the objective function
\vspace{-0.5em}
\begin{equation}
    \sum_{i=0}^{n+1} \norm{C_i-C_i'}^2,
\vspace{-0.5em}
\end{equation}
i.e., the sum of squares of distances from $n$ joints and two endpoints of the robot to the corresponding samples on the desired backbone curve. 

We then derive the constraints for the optimization problem based on the geometric structure of the robot.
For convenience we denote the $n+1$ links by vectors $\boldsymbol{j}'_{i} = C'_{i} - C'_{i-1}$ for $1\le i\le n+1$ pointing from one joint center to the next.
At each joint, the two links $\boldsymbol{j}'_{i}$ and $\boldsymbol{j}'_{i+1}$ belong to the same rotational plane $P_{i}$ ($1\le i\le n$). 
To describe the direction of the planes $P_{i}$, we associate a unit normal vector $\boldsymbol{n}_{i} \in \mathbb{R}^{3}$ to each of them, which yields a constraint
\begin{enumerate}[(i)]
    \item $\norm{\boldsymbol{n}_{i}}=1$ for $1\le i\le n$.
\end{enumerate}

The robot geometry of alternating pitch and yaw joints implies that any two consecutive rotational planes are orthogonal in $\mathbb{R}^3$, yielding
\begin{enumerate}[(i)]
    \setcounter{enumi}{1}
    \item $\boldsymbol{n}_{i-1}\cdot \boldsymbol{n}_{i} = 0$ for $2\le i\le n$.
\end{enumerate}

For $1\le i\le n$, the two links $\boldsymbol{j}'_{i}$ and $\boldsymbol{j}'_{i+1}$ both belong to the plane $P_{i}$ and the normal vector $\boldsymbol{n}_{i}$ is orthogonal to $P_{i}$. Thus $\boldsymbol{n}_{i}$ is orthogonal to both $\boldsymbol{j}'_{i}$ and $\boldsymbol{j}'_{i+1}$. Now for $2\le i\le n$, the three vectors $\boldsymbol{j}'_{i}, \boldsymbol{n}_{i}, \boldsymbol{n}_{i-1}$ are pairwise orthogonal. Therefore $\boldsymbol{j}'_{i}$ is parallel to the cross product $\boldsymbol{n}_{i-1} \times \boldsymbol{n}_{i}$.
Thus we have the constraint that
\begin{enumerate}[(i)]
    \setcounter{enumi}{2}
    \item $C'_{i} - C'_{i-1} = \boldsymbol{j}'_{i} = l\cdot \left(\boldsymbol{n}_{i-1} \times \boldsymbol{n}_{i}\right)$ for $2\le i\le n$,
\end{enumerate}
where $l$ is the length of each link. 

Suppose that we have the vectors $\boldsymbol{j}'_{i}$ fixed, then the internal shape of the robot is determined, and the value of the objective function will only depend on the choice of $C'_{i}$. For convenience, we let $d_{i} = C_{i} + (C'_{0} - C'_{i})$, then the objective function becomes:
\vspace{-0.5em}
\begin{equation*}
\begin{aligned}
    &\sum_{i=0}^{n+1} \norm{C_i-C_i'}^2 = \sum_{i=0}^{n+1}{\left[ (C'_{0}-d_{i})\cdot (C'_{0}-d_{i}) \right]} \\ 
    &= \sum_{i=0}^{n+1}{\left[C'_{0}\cdot C'_{0} - 2C'_{0}\cdot d_{i} + d_{i}\cdot d_{i} \right]} \\
    &= (n+2)C'_{0}\cdot C'_{0} - 2C'_{0}\cdot \left(\sum_{i=0}^{n+1}{d_{i}}\right) + \sum_{i=0}^{n+1}{\left(d_{i}\cdot d_{i}\right)} \\
    &= (n+2)\left(C'_{0} - \frac{\sum_{i=0}^{n+1}{d_{i}}}{n+2}\right)\cdot \left(C'_{0} - \frac{\sum_{i=0}^{n+1}{d_{i}}}{n+2}\right)\\
    &\quad+ \left[\sum_{i=0}^{n+1}{\left(d_{i}\cdot d_{i}\right)} - \frac{1}{n+2}\left(\sum_{i=0}^{n+1}{d_{i}}\right)\cdot\left(\sum_{i=0}^{n+1}{d_{i}}\right)\right].
\end{aligned}
\end{equation*}
The term in the last square bracket is a constant, therefore the objective function attains the minimum if and only if $C'_{0} = \frac{\sum_{i=0}^{n+1}{d_{i}}}{n+2}$, which is equivalent to $\sum_{i=0}^{n+1}{C'_{i}} = \sum_{i=0}^{n+1}{C_{i}}$. As a necessary mathematical constraint that ensures the objective function to attain the minimum, we have
\begin{enumerate}[(i)]
    \setcounter{enumi}{3}
    \item $\sum_{i=0}^{n+1}{C'_{i}} = \sum_{i=0}^{n+1}{C_{i}}.$
\end{enumerate}
A physical interpretation of this mathematical constraint is that the centroid of $n$ joints and two endpoints of the robot should overlap with the centroid of their corresponding  samples on the desired backbone curve.

Now we have formalized an optimization problem that exploits the geometric structure of the robot
\vspace{-0.5em}
\begin{equation*}
\begin{aligned}
& \underset{C_i'}{\text{minimize}}
& & \sum_{i=0}^{n+1} \norm{C_i-C_i'}^2 \\
& \text{subject to}
& & \text{constraints (i), (ii), (iii) and (iv)}.
\end{aligned}
\end{equation*}

The constrained nonlinear optimization problem can be solve by standard gradient-descent algorithms such as MATLAB's built-in function \texttt{fmincon} \cite{MATLAB:2020}. 
This implementation returns a local minimum of the objective function and works well in our experiments. 
In the optimization process for a continuous sequence of desired backbone curve examples, we chose zero configuration coordinates ($C_i = \boldsymbol{0}$) as the initial value for the first desired backbone curve example.
Then we used the output optimal configuration coordinates of the current desired backbone curve example as the initial seed for the next desired backbone curve example. 

\vspace{-0.5em}
\subsection{Joint angle translation}
Our optimization procedure outputs the coordinates $C'_0, C'_1, ..., C'_{n+1} \in \mathbb{R}^{3}$, which denote the optimal positions for $n$ joints, the anterior and the posterior endpoints of the robot in the work-space. We translate the coordinates to the robot joint angles in the joint space so that the result can be implemented on the physical robot. The absolute value of the joint angle is characterized by the inner product of two consecutive links at the joint. The orientation of the rotations within the plane $P_{i}$ can be specified by the direction of the normal vector $\boldsymbol{n}_{i}$, which is known after solving the optimization problem.


\begin{figure*}[t]
\centering
\vspace{0.7em}
\includegraphics[width=0.9\textwidth]{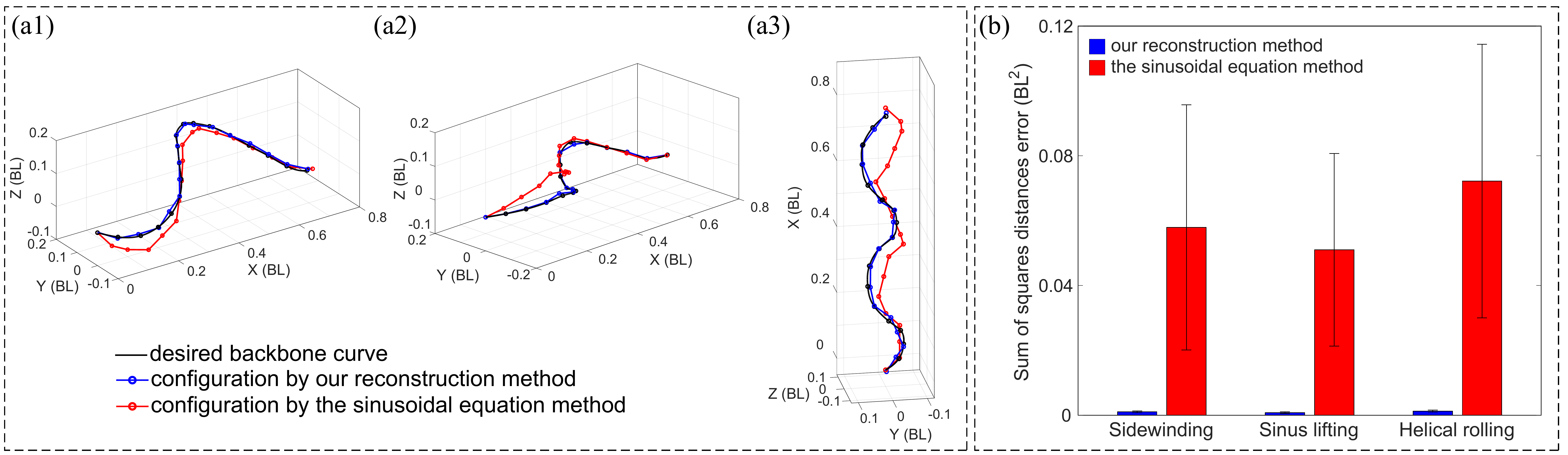}
\caption{(a) Comparisons for the robot configurations by our reconstruction method and the sinusoidal equation method, and the desired backbone curves in (a1) a sidewinding gait, (a2) a sinus lifting gait, and (a3) a helical rolling gait. All coordinates are normalized by the body length (BL) of the robot. our reconstruction method allows more accurate curve-configuration conversions. (b) A comparison of averaged discrepancies between the robot configurations and the desired backbone curves for selected sidewinding, sinus lifting and helical rolling gaits among 200 backbone curve examples on the virtual robot model. Discrepancies are evaluated by the sum of squares distances error between the positions of the robot's joints and the positions of corresponding samples on the desired backbone curve. Error bars indicate the standard deviation. Our method achieves an error between the desired and achieved curves that is an order of magnitude lower than the sinusoidal equation method.}
\vspace{-2.0em}
\label{fig:config}
\end{figure*}

\vspace{-0.5em}
\section{Results}
\subsection{Quality of backbone curve-robot configuration conversion}
We first tested the proposed 3D curve reconstruction method on the virtual twist-free alternating-rotation robot model to evaluate the quality of converting the desired backbone curve to the robot configuration.
To verify the effectiveness of the reconstruction on different types of 3D backbone curves, we tested it on the backbone curves for three widely studied 3D gaits -- sidewinding \cite{burdick1993sidewinding,astley2015modulation,zhong2020frequency}, sinus lifting \cite{tanaka2009study, ma2004dynamic}, and helical rolling \cite{zhen2015modeling,rollinson2013gait}.
As a comparison, we also implemented the approach that is commonly used in the snake robots with an alternating-rotation geometry -- prescribing the pitch and yaw joint angles with two parameterized sinusoidal equations (hereafter we refer to this approach as ``the sinusoidal equation method"), as in \cite{astley2015modulation, zhong2020frequency, tesch2009parameterized, hatton2010generating} discussed in Section \ref{sec:related_work}.
We measured the discrepancies between the robot configurations by our reconstruction method and the desired curves and the discrepancies between the robot configurations by the sinusoidal equation method and the desired curves.
The discrepancy between the robot configuration and the desired curve is evaluated by the sum of squares distances error criterion $D = \sum_{i=0}^{n+1} \norm{C_i-C_i'}^2$, where $C_i$ are the positions for desired backbone curve samples and $C_i'$ are the positions of the joints and two endpoints of the robot.

\subsubsection{Sidewinding}\label{sec:sidewinding}
The desired finite-length continuous backbone curves in 3D Cartesian space for sidewinding were achieved with parametric equations.
\vspace{-0.5em}
\begin{equation}
\begin{aligned}
    y &= A_y\sin(\omega_y x + f t)\\
    z &= A_z\sin(\omega_z x + f t + \phi),
    \label{eq:curve_equation}
\vspace{-0.5em}
\end{aligned}
\end{equation}
where we fixed $x \in [0,1]$, $A_y = \pi/4$, $A_z = \pi/3$, $\omega_y = \omega_z = 2\pi$, $f = 1$, and $\phi = -\pi/2$. 
A sequence of 200 backbone curve examples in a gait cycle were then achieved by varying $t$.
A comparison of the robot configuration by our reconstruction method, the robot configuration by the sinusoidal equation method, and the desired backbone curve is shown in Fig. \ref{fig:config}(a1), the coordinates are normalized by the body length of the robot (BL). 
For the 200 backbone curve examples, the averaged sum of squares distances error for the robot configurations by our reconstruction method is $0.0010 \pm 0.0003$ BL$^2$, while the same measure for the sinusoidal equation method is $0.0579 \pm 0.0378$ BL$^2$, as illustrated in Fig. \ref{fig:config}(b).

\subsubsection{Sinus lifting}\label{sec:sinus_lifting}
We achieved sinus lifting backbone curves by setting the parameters in \eqref{eq:curve_equation} as $x \in [0,1]$, $A_y = A_z = \pi/4$, $\omega_y = 2\pi$, $\omega_z = 3\pi$, $f = 1$, and $\phi = \pi/2$. 
Similar to the method described in the previous section, we performed comparisons over 200 backbone curve examples, one of which is as depicted in Fig. \ref{fig:config}(a2). 
The average of the sum of squares distances error for the robot configurations by our reconstruction method is $0.0007 \pm 0.0002$ BL$^2$, which is much smaller than the same measure for the sinusoidal equation method $0.0510 \pm 0.0297$ BL$^2$, as shown in Fig. \ref{fig:config}(b).

\subsubsection{Helical rolling}
Backbone curves for a helical rolling gait can also be achieved with parameters $x \in [0,1]$, $A_x = A_y = \pi/3$, $\omega_x = \omega_y = 4\pi$, $\phi = \pi/2$, and $f = 1$ in \eqref{eq:curve_equation}. 
A sequence of 200 desired backbone curve examples covering a full gait cycle were collected by varying $t$ in the same manner used in previous sections.
The comparison between an example of the desired curve and the robot configurations is demonstrated in Fig. \ref{fig:config}(a3).
The robot configurations by our reconstruction method achieve a $0.0012 \pm 0.0003$ BL$^2$ averaged sum of squares distances error to the desired curves over 200 examples, compared to $0.0747 \pm 0.0422$ BL$^2$ obtained by the sinusoidal equation method, as shown in Fig. \ref{fig:config}(b). 

Larger discrepancies in the sinusoidal equation method suggest that although the sinusoidal equation method replicates the projection of the desired 3D backbone curve onto the horizontal and vertical planes, but does not necessarily result in an accurate approximation of the full 3D backbone curve.
By directly solving the optimization problem in the full 3D space, our reconstruction method achieves significantly smaller discrepancies between the output and desired backbone curves than does the sinusoidal equation method, such that the robot body can better approximate the desired 3D shape.

\vspace{-0.5em}
\subsection{Real-time implementation in physical robot locomotion}
We performed the reconstruction method in real-time to create joint angle set-points for a snake robot. 
In the following robot experiments, a snake robot composed of sixteen identical actuated one DOF bending joints was used, as shown by Fig.~\ref{fig:robot}.
The joints are arranged such that the neighboring modules' axes of rotation are torsionally rotated ninety degrees relative to each other, yielding an alternating-pitch-and-yaw twist-free geometry.
In each joint, a low-level PID controller is embedded, which controls the actuators to follow the joint angle set points. 
Experiments were conducted on flat, smooth, hard ground, where we assume the ground reaction forces are given by kinetic Coulomb friction.
For each experiment, we conducted five trials for each gait tested, and we commanded the snake robot to execute two full gait cycles for each trial. 
We tracked the snake robot's trajectory via eight reflective markers attached evenly along the backbone of the robot and an OptiTrack motion capture system.

To test the effectiveness of our reconstruction method in real-time, we conducted experiments of the robot executing a series of sidewinding gaits with varying temporal frequencies (gait speeds) with the our reconstruction method, and with the sinusoidal equation method, respectively.
Following the gait design reported in \cite{zhong2020frequency}, desired 3D backbone curves were composed by sinusoidal waves in horizontal plane and sigmoid-filtered sinusoidal waves in vertical plane, prescribed by
\vspace{-0.5em}
\begin{equation}
\begin{aligned}
    y &= A_y\sin(\omega_y x + f t)\\
    z &= A_z \sigma[\sin(\omega_z x + f t + \phi)],
    \label{eq:curve_equation_3}
\vspace{-0.5em}
\end{aligned}
\end{equation}
where $\sigma(t) = \frac{1}{1+e^{-\gamma t}}$. 
We fixed $\gamma = 4$, $A_y = \pi/4$, $A_z = \pi/3$, $\omega_y = \omega_z = 2\pi$, and $\phi = -\pi/2$.
We varied the temporal frequency $f \in (0,2]$ Hz to achieve a family of sidewinding gaits with different gait speeds to test the real-time performance of our reconstruction method. 
The expected motion for these gaits is pure translation without body rotation.

\begin{figure}[t]
\centering
\vspace{0.7em}
\includegraphics[width=0.98\columnwidth]{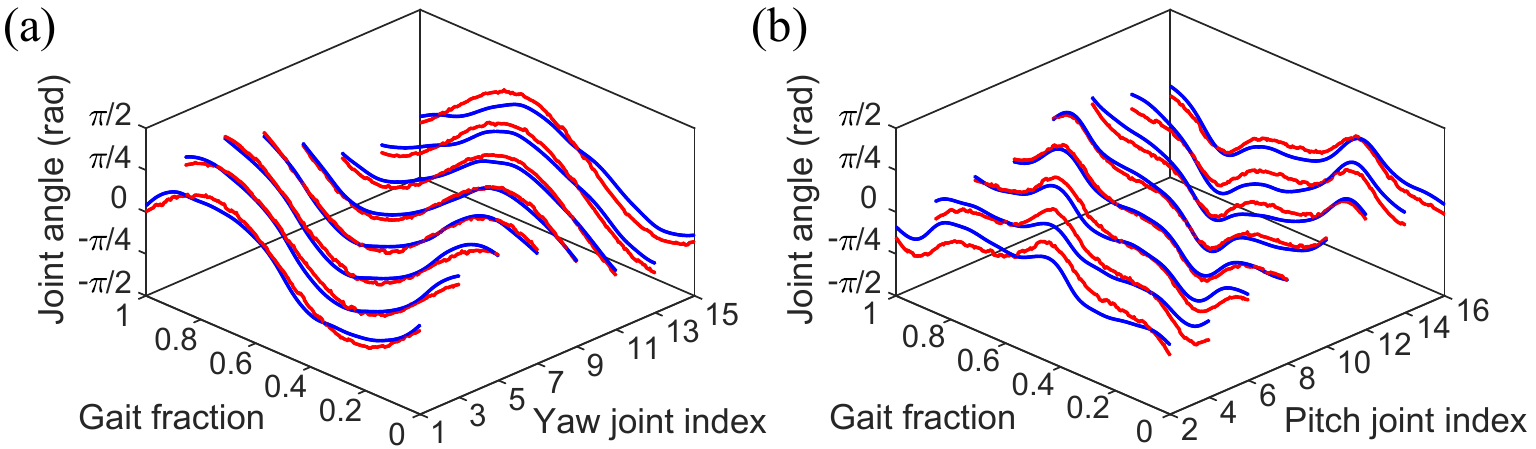}
\caption{An example comparison of output joint angle trajectories for the same sequence of backbone curves by different optimization algorithms for the sidewinding gait with $f=1$ Hz. Red trajectories are achieved by running the sinusoidal equation method (implementing the ``annealed chain fitting" optimization algorithm introduced in \cite{hatton2010generating}). Blue trajectories are achieved by running the proposed optimization algorithm in this work. The joint angle trajectories output by the method we proposed in this work are smooth, which are able to be used directly on the robot to produce a continuous locomotion, while the joint angle trajectories by the sinusoidal equation method are not smooth, which need a post-processing step of smoothing.}
\vspace{-2.0em}
\label{fig:joint_angle}
\end{figure}

\begin{figure}[t]
\centering
\vspace{0.7em}
\includegraphics[width=0.8\columnwidth]{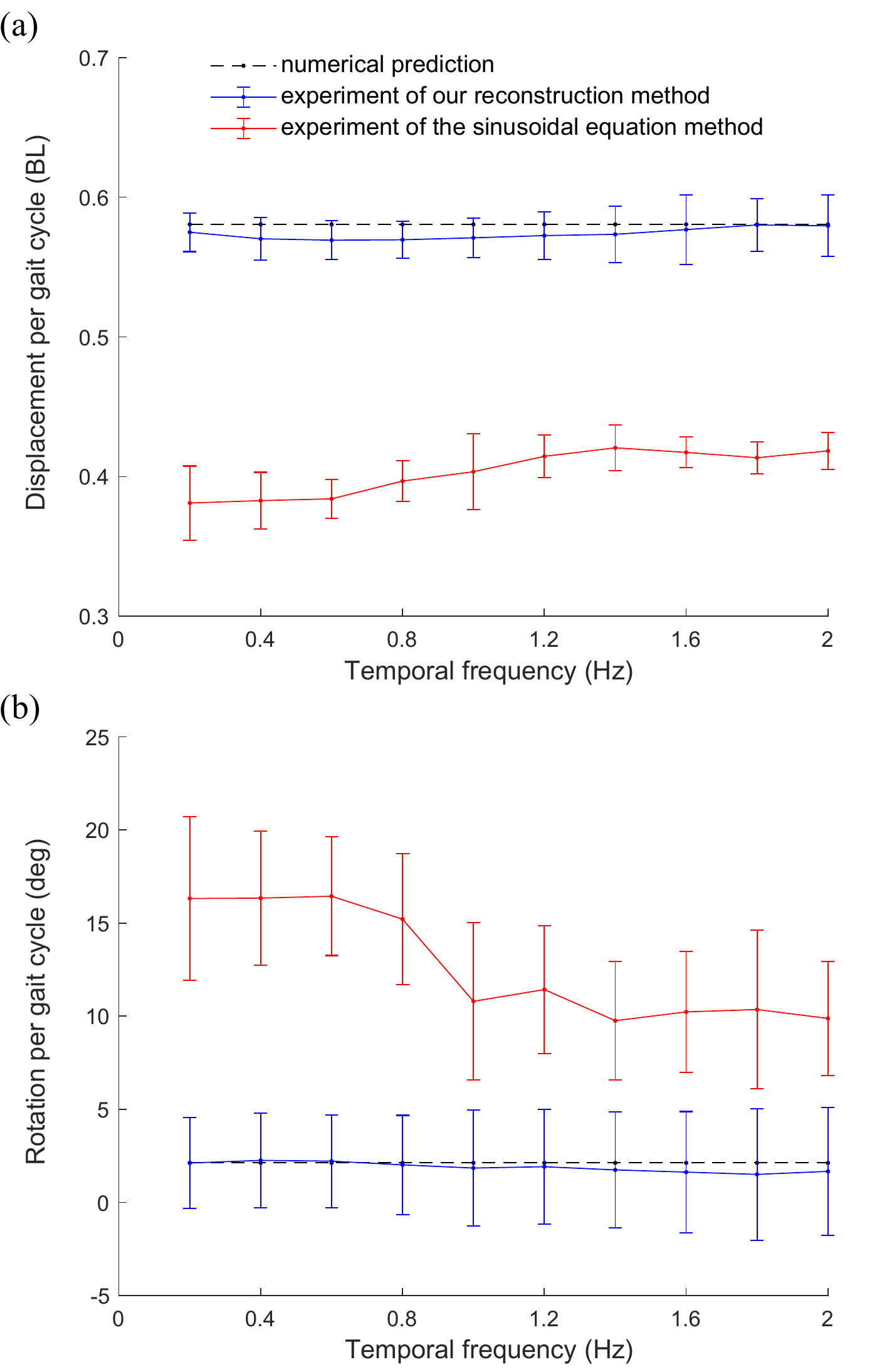}
\caption{The averaged body displacement and body rotation of the robot over a gait cycle for the sidewinding gaits with different temporal frequencies. The error bars are standard deviations.}
\vspace{-1.0em}
\label{fig:performance}
\end{figure}

To verify that the joint angle sequence output by our optimization algorithm can produce a continuous locomotion of the robot, we first examined the smoothness of joint angle trajectories and compared them with the joint angle trajectories output by the sinusoidal equation method (implementing an optimization algorithm \textit{annealed chain fitting} introduced in \cite{hatton2010generating}) given the same sequence of backbone curves. 
Fig. \ref{fig:joint_angle} visualized the example joint angle trajectories by two methods.
We calculated the averaged difference between neighbouring joint angles over the chronological sequence of 200 robot configurations in a full gait cycle.
Taking the gait with $f=1$ Hz as an example, we found that the sinusoidal equation method resulted in an average joint angle change between time steps of 10.4 degrees, whereas our reconstruction method resulted in only 1.6 degrees change between time steps. 
The numerical comparisons of the smoothness for all tested gaits see Table \ref{table:1}.
This comparison demonstrated the joint angle trajectories output by the sinusoidal equation method are not smooth, so that they require a post-processing smoothing step.
The trajectories output by our reconstruction method do not need a smoothing step and can be used directly on the robot to produce a continuous motion.

\begin{figure}[t]
\centering
\vspace{0.7em}
\includegraphics[width=0.9\columnwidth]{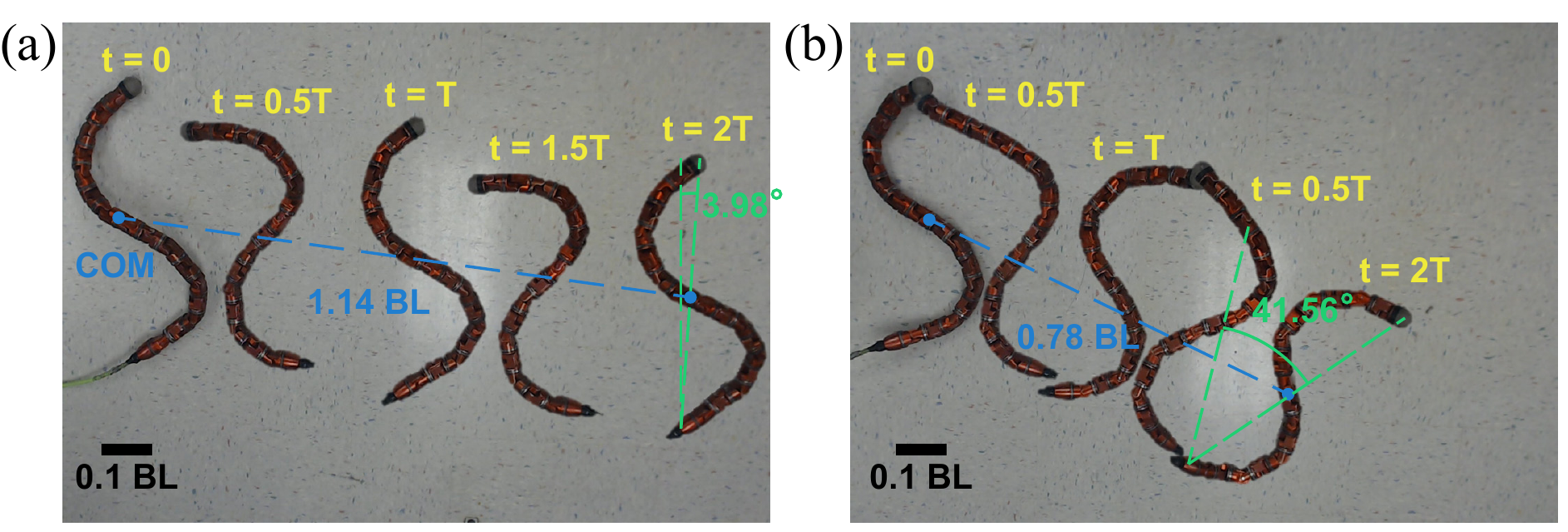}
\caption{Example video frame sequences of the robot experiments of the sidewinding gait at $f = 1$ Hz (a) with our reconstruction method, and (b) with the sinusoidal equation method. The desired motion for the robot is to translate from left to right without body rotation. Here time $t$ is measured in terms of complete gait cycle period $T$.}
\vspace{-1.9em}
\label{fig:robot_frames}
\end{figure}

We then studied the robot's locomotion by measuring the robot's body displacement and body rotation.
We compared the motion of the gaits by our reconstruction method and the gaits by the sinusoidal equation method with the desired motion from a numerical prediction.
In the numerical prediction, the contact state pattern is prescribed with the assumption that the robots' configuration is a 3D continuous curve \cite{astley2020surprising}. 
The displacement is calculated by numerically integrating the equations of motion for the continuous curve throughout one period based on the resistive force theory \cite{hatton2015nonconservativity,li2013terradynamics}, which has been demonstrated as a reliable method to provide predictions for snake robot locomotion \cite{astley2015modulation,wang2020omega,chong2018coordination}.
Fig. \ref{fig:performance} shows the quantitative results of the robot's body displacement and body rotation over a gait cycle versus temporal frequency $f$. 
We found that the robot's body displacement and body rotation with our method agree with the numerical predictions, which indicates that our method allows the discrete robot to approach the performance of a desired motion of a sequence of desired continuous backbone curves.
In contrast, the robot's body displacement of gaits by the sinusoidal equation method cannot reach the predictions, and large unexpected body rotation emerges during the gait cycle.
The results also demonstrated that the reconstruction method enables the robot to perform robustly over different temporal frequencies (gait speeds) as predicted, while the robot's locomotion with the sinusoidal equation method varies as temporal frequency changes. 
We reported numerical results in Table \ref{table:1}.
Example video frame sequences of the robot experiments of executing the gaits by our reconstruction method and by the sinusoidal equation method with $f=1$ Hz are shown in Fig. \ref{fig:robot_frames}, which illustrates that the gait by our reconstruction method generates larger body displacement and less body rotation. 

\begin{figure}[t]
\centering
\vspace{0.7em}
\includegraphics[width=0.98\columnwidth]{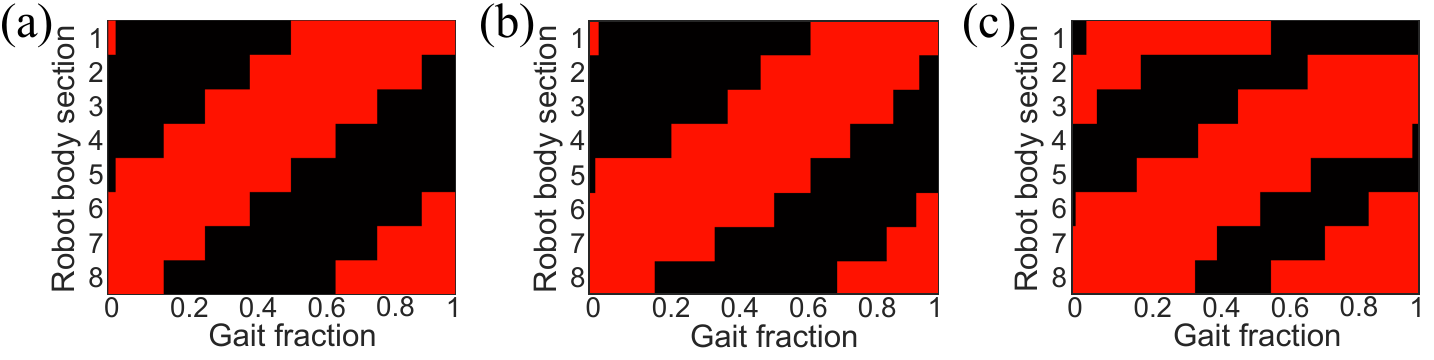}
\caption{An example comparison of contact state patterns for the sidewinding gait at $f=1$ Hz: (a) the desired contact state pattern of the robot, (b) the contact state pattern of the robot executing the gait by our reconstruction method, and (c) the contact state pattern of the robot executing the gait by the sinusoidal equation method. The contact state pattern of the gait by our reconstruction method shares a higher similarity (89.4\%) with the desired contact state pattern than the gait by the sinusoidal equation method (57.2\%).}
\vspace{-1.4em}
\label{fig:contact}
\end{figure}

To further investigate how our reconstruction method impacts the robot's locomotion, we compared the robot's contact state patterns during a gait cycle with the desired contact state pattern that generated by the numerical prediction of the desired gait. 
We obtained the contact state patterns for the gaits with our reconstruction method and the gaits with the sinusoidal equation method from the tracked motion data. 
At each time step, if a marker's height lied below the average of height of the tracking markers, we considered its corresponding body section to be in contact with the ground; otherwise, we considered the body section of the robot to be not in contact.
We found that the contact state patterns for the gaits by our reconstruction method share a high similarity with the desired contact state patterns while the contact state patterns by the sinusoidal equation method not. 
The comparisons of similarity percentages for all the gaits we tested are reported in Table \ref{table:1}.
For example, for the gait with $f=1$ Hz, our method yields an averaged 89.4\% overlap, while the number for the gaits by the sinusoidal equation method is 57.2\%, the contact state patterns for this gait are illustrated in Fig. \ref{fig:contact}.
This comparison indicates that the reconstruction method allows more accurate conversions from the desired backbone curves to robot configurations.
Therefore, the robot can achieve a contact state pattern that shares a higher similarity with the desired pattern, which enables the robot to achieve the desired locomotion of a sequence of desired continuous backbone curves. 

\begin{table}[t]\footnotesize
\centering
\vspace{0.7em}
\begin{tabular}{ |c|c|c|c|c|c| } 
\hline
\multicolumn{2}{|c|}{Gait} & \multirow{2}*{S (deg)} & \multirow{2}*{D (BL)} & \multirow{2}*{R (deg)} & \multirow{2}*{CSPS (\%)} \\
\cline{1-2}
Method & $f$ (Hz) & ~ & ~ & ~ & ~\\
\hline
\multirow{10}*{\makecell[c]{Our\\reconstruction\\method}} & 0.2 & 0.38 & 0.575 & 2.12 & 91.5\\
\cline{2-6}
~ & 0.4 & 0.62 & 0.570 & 2.26 & 91.1\\
\cline{2-6}
~ & 0.6 & 0.99 & 0.569 & 2.22 & 89.7\\
\cline{2-6}
~ & 0.8 & 1.27 & 0.570 & 2.02 & 90.2\\
\cline{2-6}
~ & 1.0 & 1.64 & 0.571 & 1.85 & 89.4\\
\cline{2-6}
~ & 1.2 & 2.01 & 0.573 & 1.92 & 88.2\\
\cline{2-6}
~ & 1.4 & 2.52 & 0.573 & 1.75 & 86.5\\
\cline{2-6}
~ & 1.6 & 3.14 & 0.577 & 1.63 & 88.1\\
\cline{2-6}
~ & 1.8 & 3.79 & 0.580 & 1.51 & 86.8\\
\cline{2-6}
~ & 2.0 & 4.53 & 0.580 & 1.67 & 87.3\\
\hline
\multirow{10}*{\makecell[c]{The\\sinusoidal\\equation\\method}} & 0.2 & 2.72 & 0.381 & 16.32 & 70.9\\
\cline{2-6}
~ & 0.4 & 3.91 & 0.383 & 16.34 & 65.5\\
\cline{2-6}
~ & 0.6 & 5.50 & 0.384 & 16.44 & 63.3\\
\cline{2-6}
~ & 0.8 & 8.03 & 0.397 & 15.21 & 65.8\\
\cline{2-6}
~ & 1.0 & 10.42 & 0.404 & 10.80 & 57.2\\
\cline{2-6}
~ & 1.2 & 13.29 & 0.415 & 11.43 & 60.0\\
\cline{2-6}
~ & 1.4 & 15.86 & 0.421 & 9.76 & 52.0\\
\cline{2-6}
~ & 1.6 & 18.58 & 0.417 & 10.23 & 51.7\\
\cline{2-6}
~ & 1.8 & 20.90 & 0.414 & 10.36 & 46.4\\
\cline{2-6}
~ & 2.0 & 24.11 & 0.418 & 9.88 & 48.9\\
\hline
\end{tabular}
\caption{Robot performance comparisons of the gaits family (as in \eqref{eq:curve_equation_3}, expected to generate pure translation without rotation) by our reconstruction method and by the sinusoidal equation method with metrics of smoothness (average joint angle change between timesteps, denoted as S, lower is better), averaged displacement per gait cycle (D, higher is better), body rotation per gait cycle (R, lower is better), and contact state pattern similarity (CSPS, higher is better) with the desired pattern. Over the full range of frequencies, our method performed better than the sinusoidal equation method by all of these metrics.}
\label{table:1}
\vspace{-3.1em}
\end{table}
\vspace{-0.2em}
\section{Conclusion}
In this paper, we presented a reconstruction method that can realize 3D continuous backbone curve on discrete snake robots with a twist-free alternating pitch and yaw joints design.
Our analysis showed that our reconstruction method allows more accurate curve-configuration conversions than the state-of-the-art when applied to three types of gaits widely used in snake robot locomotion. 
We also conducted physical robot experiments to compare our reconstruction method and the sinusoidal equation method, in which the robot was commanded to execute a family of sidewinding gaits with different temporal frequencies.
We experimentally validated our optimization algorithm enables that small changes in the desired backbone curves only result in small changes in the robot configuration, which lead to continuous robot locomotion.
By comparing the body displacement and rotation, we found that the gaits from our reconstruction method outperformed the gait by the sinusoidal equation method, generating larger body displacement and less undesired body rotation.
Furthermore, we verified the robot's locomotive performance and contact state pattern, when executing the gaits from our reconstruction method, match with the numerical prediction of the desired motion generated by a sequence of desired continuous backbone curves, while the gaits by the sinusoidal equation method do not.
Our reconstruction method produces more accurate conversions from the desired continuous backbone curves to discrete robot configurations.
Thus, this method allows twist-free snake robots with robust, simple, single-axis rotary joints to locomote using a sequence of continuous backbone curves.

The proposed reconstruction method focuses on the specific snake robot design of alternating pitch and yaw joints.
Future work will investigate the extension of this idea to the applications on other twist-free mobile limbless robots and continuum manipulators.
Further, we hope to integrate the reconstruction method into the shape-based compliant control framework for snake robot locomotion \cite{travers2016shape,wang2020directional}, where it can help the robot comply to challenging 3D terrains and irregular 3D obstacles.


\bibliographystyle{IEEEtran}

\bibliography{RAL2021_reconstruction}

\begin{thebibliography}{10}
\providecommand{\url}[1]{#1}
\csname url@rmstyle\endcsname
\providecommand{\newblock}{\relax}
\providecommand{\bibinfo}[2]{#2}
\providecommand\BIBentrySTDinterwordspacing{\spaceskip=0pt\relax}
\providecommand\BIBentryALTinterwordstretchfactor{4}
\providecommand\BIBentryALTinterwordspacing{\spaceskip=\fontdimen2\font plus
\BIBentryALTinterwordstretchfactor\fontdimen3\font minus
  \fontdimen4\font\relax}
\providecommand\BIBforeignlanguage[2]{{%
\expandafter\ifx\csname l@#1\endcsname\relax
\typeout{** WARNING: IEEEtran.bst: No hyphenation pattern has been}%
\typeout{** loaded for the language `#1'. Using the pattern for}%
\typeout{** the default language instead.}%
\else
\language=\csname l@#1\endcsname
\fi
#2}}

\bibitem{wright2007design}
C.~Wright, A.~Johnson, A.~Peck, Z.~McCord, A.~Naaktgeboren, P.~Gianfortoni,
  M.~Gonzalez-Rivero, R.~Hatton, and H.~Choset, ``Design of a modular snake
  robot,'' in \emph{2007 IEEE/RSJ International Conference on Intelligent
  Robots and Systems}.\hskip 1em plus 0.5em minus 0.4em\relax IEEE, 2007, pp.
  2609--2614.

\bibitem{takaoka2011snake}
S.~Takaoka, H.~Yamada, and S.~Hirose, ``Snake-like active wheel robot acm-r4. 1
  with joint torque sensor and limiter,'' in \emph{2011 IEEE/RSJ International
  Conference on Intelligent Robots and Systems}.\hskip 1em plus 0.5em minus
  0.4em\relax IEEE, 2011, pp. 1081--1086.

\bibitem{transeth2008snake}
A.~A. Transeth, R.~I. Leine, C.~Glocker, K.~Y. Pettersen, and P.~Liljeb{\"a}ck,
  ``Snake robot obstacle-aided locomotion: Modeling, simulations, and
  experiments,'' \emph{IEEE Transactions on Robotics}, vol.~24, no.~1, pp.
  88--104, 2008.

\bibitem{fu2020lateral}
Q.~Fu, S.~W. Gart, T.~W. Mitchel, J.~S. Kim, G.~S. Chirikjian, and C.~Li,
  ``Lateral oscillation and body compliance help snakes and snake robots stably
  traverse large, smooth obstacles,'' \emph{Integrative and Comparative
  Biology}, 2020.

\bibitem{nilsson1998snake}
M.~Nilsson, ``Why snake robots need torsion-free joints and how to design
  them,'' in \emph{Proceedings. 1998 IEEE International Conference on Robotics
  and Automation (Cat. No. 98CH36146)}, vol.~1.\hskip 1em plus 0.5em minus
  0.4em\relax IEEE, 1998, pp. 412--417.

\bibitem{hirose1993biologically}
S.~Hirose, ``Biologically inspired robots,'' \emph{Snake-Like Locomotors and
  Manipulators}, 1993.

\bibitem{chirikjian1994modal}
G.~S. Chirikjian and J.~W. Burdick, ``A modal approach to hyper-redundant
  manipulator kinematics,'' \emph{IEEE Transactions on Robotics and
  Automation}, vol.~10, no.~3, pp. 343--354, 1994.

\bibitem{hatton2010generating}
R.~L. Hatton and H.~Choset, ``Generating gaits for snake robots: annealed chain
  fitting and keyframe wave extraction,'' \emph{Autonomous Robots}, vol.~28,
  no.~3, pp. 271--281, 2010.

\bibitem{takemori2018gait}
T.~Takemori, M.~Tanaka, and F.~Matsuno, ``Gait design for a snake robot by
  connecting curve segments and experimental demonstration,'' \emph{IEEE
  Transactions on Robotics}, vol.~34, no.~5, pp. 1384--1391, 2018.

\bibitem{burdick1993sidewinding}
J.~W. Burdick, J.~Radford, and G.~S. Chirikjian, ``A'sidewinding'locomotion
  gait for hyper-redundant robots,'' in \emph{[1993] Proceedings IEEE
  International Conference on Robotics and Automation}.\hskip 1em plus 0.5em
  minus 0.4em\relax IEEE, 1993, pp. 101--106.

\bibitem{ma2004dynamic}
S.~Ma, Y.~Ohmameuda, and K.~Inoue, ``Dynamic analysis of 3-dimensional snake
  robots,'' in \emph{2004 IEEE/RSJ International Conference on Intelligent
  Robots and Systems (IROS)(IEEE Cat. No. 04CH37566)}, vol.~1.\hskip 1em plus
  0.5em minus 0.4em\relax IEEE, 2004, pp. 767--772.

\bibitem{tesch2009parameterized}
M.~Tesch, K.~Lipkin, I.~Brown, R.~Hatton, A.~Peck, J.~Rembisz, and H.~Choset,
  ``Parameterized and scripted gaits for modular snake robots,'' \emph{Advanced
  Robotics}, vol.~23, no.~9, pp. 1131--1158, 2009.

\bibitem{astley2015modulation}
H.~C. Astley, C.~Gong, J.~Dai, M.~Travers, M.~M. Serrano, P.~A. Vela,
  H.~Choset, J.~R. Mendelson, D.~L. Hu, and D.~I. Goldman, ``Modulation of
  orthogonal body waves enables high maneuverability in sidewinding
  locomotion,'' \emph{Proceedings of the National Academy of Sciences}, vol.
  112, no.~19, pp. 6200--6205, 2015.

\bibitem{zhong2020frequency}
B.~Zhong, T.~Wang, J.~Rieser, A.~Kaba, H.~Choset, and D.~Goldman, ``Frequency
  modulation of body waves to improve performance of limbless robots,'' in
  \emph{Robotics: Science and Systems}, 2020.

\bibitem{o2014elementary}
B.~O'neill, \emph{Elementary differential geometry}.\hskip 1em plus 0.5em minus
  0.4em\relax Academic press, 2014.

\bibitem{chirikjian1995kinematics}
G.~S. Chirikjian and J.~W. Burdick, ``The kinematics of hyper-redundant robot
  locomotion,'' \emph{IEEE transactions on robotics and automation}, vol.~11,
  no.~6, pp. 781--793, 1995.

\bibitem{mochiyama1999shape}
H.~Mochiyama, E.~Shimemura, and H.~Kobayashi, ``Shape control of manipulators
  with hyper degrees of freedom,'' \emph{The International Journal of Robotics
  Research}, vol.~18, no.~6, pp. 584--600, 1999.

\bibitem{andersson2008discretization}
S.~B. Andersson, ``Discretization of a continuous curve,'' \emph{IEEE
  Transactions on Robotics}, vol.~24, no.~2, pp. 456--461, 2008.

\bibitem{tanaka2009study}
M.~Tanaka and F.~Matsuno, ``A study on sinus-lifting motion of a snake robot
  with switching constraints,'' in \emph{2009 IEEE International Conference on
  Robotics and Automation}.\hskip 1em plus 0.5em minus 0.4em\relax IEEE, 2009,
  pp. 2270--2275.

\bibitem{zhen2015modeling}
W.~Zhen, C.~Gong, and H.~Choset, ``Modeling rolling gaits of a snake robot,''
  in \emph{2015 IEEE International Conference on Robotics and Automation
  (ICRA)}.\hskip 1em plus 0.5em minus 0.4em\relax IEEE, 2015, pp. 3741--3746.

\bibitem{MATLAB:2020}
MATLAB, \emph{version 9.8.0 (R2020a)}.\hskip 1em plus 0.5em minus 0.4em\relax
  Natick, Massachusetts: The MathWorks Inc., 2020.

\bibitem{rollinson2013gait}
D.~Rollinson and H.~Choset, ``Gait-based compliant control for snake robots,''
  in \emph{2013 IEEE International Conference on Robotics and
  Automation}.\hskip 1em plus 0.5em minus 0.4em\relax IEEE, 2013, pp.
  5138--5143.

\bibitem{astley2020surprising}
H.~C. Astley, J.~R. Mendelson, J.~Dai, C.~Gong, B.~Chong, J.~M. Rieser, P.~E.
  Schiebel, S.~S. Sharpe, R.~L. Hatton, H.~Choset, \emph{et~al.}, ``Surprising
  simplicities and syntheses in limbless self-propulsion in sand,''
  \emph{Journal of Experimental Biology}, vol. 223, no.~5, 2020.

\bibitem{hatton2015nonconservativity}
R.~L. Hatton and H.~Choset, ``Nonconservativity and noncommutativity in
  locomotion,'' \emph{The European Physical Journal Special Topics}, vol. 224,
  no.~17, pp. 3141--3174, 2015.

\bibitem{li2013terradynamics}
C.~Li, T.~Zhang, and D.~I. Goldman, ``A terradynamics of legged locomotion on
  granular media,'' \emph{science}, vol. 339, no. 6126, pp. 1408--1412, 2013.

\bibitem{wang2020omega}
T.~Wang, B.~Chong, K.~Diaz, J.~Whitman, H.~Lu, M.~Travers, D.~I. Goldman, and
  H.~Choset, ``The omega turn: a biologically-inspired turning strategy for
  elongated limbless robots,'' in \emph{2020 IEEE/RSJ International Conference
  on Intelligent Robots and Systems (IROS)}.\hskip 1em plus 0.5em minus
  0.4em\relax IEEE, 2020, pp. 000--000.

\bibitem{chong2018coordination}
B.~Chong, Y.~O. Aydin, C.~Gong, G.~Sartoretti, Y.~Wu, J.~M. Rieser, H.~Xing,
  J.~W. Rankin, K.~Michel, A.~G. Nicieza, \emph{et~al.}, ``Coordination of back
  bending and leg movements for quadrupedal locomotion.'' in \emph{Robotics:
  Science and Systems}, 2018.

\bibitem{travers2016shape}
M.~J. Travers, J.~Whitman, P.~E. Schiebel, D.~I. Goldman, and H.~Choset,
  ``Shape-based compliance in locomotion.'' in \emph{Robotics: Science and
  Systems}, 2016.

\bibitem{wang2020directional}
T.~{Wang}, J.~{Whitman}, M.~{Travers}, and H.~{Choset}, ``Directional
  compliance in obstacle-aided navigation for snake robots,'' in \emph{2020
  American Control Conference (ACC)}, 2020, pp. 2458--2463.

\end{thebibliography}

\end{document}